\documentclass[preprint,nonatbib]{elsarticle}
\usepackage{xpatch}
\usepackage{amsmath}
\usepackage{graphicx}
\usepackage[backend=biber,sorting=none,maxnames=3,maxbibnames=99,url=false]{biblatex}
\AtEveryBibitem{\clearfield{urlyear}\clearfield{urlmonth}\clearfield{urlday}}
\usepackage{svg}
\usepackage{float}
\usepackage{chngcntr}
\usepackage{etoolbox}
\usepackage{booktabs}
\usepackage[a4paper, total={6in, 8in}]{geometry}
\usepackage{lineno}
\usepackage{algorithm}
\usepackage{algpseudocode}
\floatname{algorithm}{Stage}
\usepackage{xcolor}
\usepackage{listings}
\nolinenumbers

\lstdefinestyle{promptstyle}{
    basicstyle=\small\ttfamily\color{blue!60!black},
    breaklines=true,
    breakatwhitespace=true,
    frame=single,
    framesep=5pt,
    rulecolor=\color{black!50},
    backgroundcolor=\color{white},
    xleftmargin=5pt,
    xrightmargin=5pt,
    columns=fullflexible,
    keepspaces=true,
    showstringspaces=false
}

\title{Agentic AI Home Energy Management System: A Large Language Model Framework for Residential Load Scheduling}

\author[eeg]{Reda El Makroum}
\ead{elmakroum@eeg.tuwien.ac.at}

\author[eeg,ntnu]{Sebastian Zwickl-Bernhard}
\ead{zwickl@eeg.tuwien.ac.at}

\author[eeg]{Lukas Kranzl}
\ead{kranzl@eeg.tuwien.ac.at}

\address[eeg]{Energy Economics Group (EEG), Technische Universität Wien, Gußhausstraße 25-29/E370-3, 1040 Vienna, Austria}
\address[ntnu]{Department of Industrial Economics and Technology Management, The Norwegian University of Science and Technology, Trondheim, Norway}

\makeatletter
\def\ps@pprintTitle{}
\makeatother

\xpatchcmd{\pprintMaketitle}{\hrule}{}{}{}
\xpatchcmd{\pprintMaketitle}{\hrule}{}{}{}

\begin{document}

\maketitle
\vspace{-6em}
\section*{Abstract}
The sustainable transformation of the electricity sector requires substantial increases in residential demand response capacity. Yet, Home Energy Management Systems (HEMS) adoption remains limited by user interaction barriers that require translating everyday preferences into technical parameters. While large language models have been applied to energy systems as code generators and parameter extractors for HEMS configuration, no existing implementation deploys LLMs as autonomous coordinators managing the complete workflow from natural language input to multi-appliance scheduling and device control. This paper presents an agentic AI HEMS where LLMs autonomously coordinate multi-appliance scheduling from natural language requests through to device control, achieving optimal scheduling without example demonstrations or few-shot learning. A hierarchical architecture combining one orchestrator agent with three specialist agents is developed using the ReAct pattern for iterative reasoning and acting, enabling dynamic coordination without hardcoded workflows while integrating Google Calendar for context-aware deadline extraction. Evaluation across three open-source models (Llama-3.3-70B, Qwen-3-32B, GPT-OSS-120B) using real Austrian day-ahead electricity prices reveals substantial capability differences for multi-appliance coordination. Llama-3.3-70B successfully coordinates all three appliances across all evaluation scenarios to match cost-optimal benchmarks computed via mixed-integer linear programming, while Qwen-3-32B and GPT-OSS-120B struggle to coordinate all appliances simultaneously despite achieving perfect single-appliance performance. Progressive prompt engineering experiments demonstrate that analytical query handling without explicit guidance remains unreliable despite models' general reasoning capabilities, requiring workflow instructions for consistent tool selection. The demonstrated system enables natural language based scheduling without technical parameter specification, offering a pathway to address configuration complexity barriers that currently limit residential HEMS adoption. All system components, including complete agent prompts, orchestration logic, and simulation user interfaces, are released as open source to enable reproducibility and further development.

\textbf{Keywords:} Agentic AI, Home Energy Management Systems, Large Language Models, Demand Response, Load Scheduling

\section{Introduction}

\subsection{Setting the Stage}

The global energy transition requires fundamental changes in how electricity is generated, distributed, and consumed across all sectors. According to the International Energy Agency, electricity grids must integrate growing shares of variable renewable generation while maintaining reliability and affordability, creating unprecedented demand for system flexibility \cite{iea_electricity_2023}. This transformation affects all sectors, with flexibility resources needed to balance supply and demand in real time. The agency further projects that global demand response capacity must reach 500 GW by 2030, representing a tenfold increase from 2020 levels, with buildings and residential electric vehicles (EVs) accounting for approximately 60\% of this potential \cite{iea_demand_2023}. The residential sector thus emerges as a critical resource due to its scale and diversity of controllable loads. Many residential loads offer temporal scheduling freedom, allowing consumption to shift across time without compromising user requirements. When coordinated effectively, these flexible appliances and systems represent significant potential for grid services, demand response participation, and renewable energy integration \cite{golmohamadi_demandside_2024}.

Realizing this potential requires systems that coordinate load scheduling to minimize electricity costs while respecting user constraints. Home Energy Management Systems (HEMS) address this need through software platforms that optimize household energy consumption via intelligent scheduling and control of appliances and distributed energy resources \cite{nebey_recent_2024}. These systems have demonstrated technical effectiveness in reducing electricity costs and enabling demand response participation \cite{liu_optimization_2020}. However, widespread residential adoption remains limited. The IEA projects that meeting global decarbonization targets requires HEMS deployments to increase from 4 million units in 2020 to 32.7 million units by 2030, yet current adoption trajectories fall far short of this eightfold growth requirement \cite{iea_demand_2023}. While financial barriers and system complexity contribute to this gap \cite{parrish_demand_2019}, user interaction challenges emerge as a critical bottleneck \cite{khafiso_barriers_2024}. Existing HEMS interfaces require users to translate everyday preferences into numerous well-formatted technical parameters, interpret often-confusing parameter documentation, and provide precise inputs for optimization algorithms. This process is time-consuming and demotivates use among non-expert users, particularly the elderly and those with limited technical literacy, thus diminishing HEMS democratization through steep learning curves \cite{michelon_large_2025}. Addressing this interaction barrier through intuitive, conversational interfaces represents a key enabler for widespread HEMS deployment.

Natural language interfaces offer promise for addressing these interaction challenges. Recent advances in large language model (LLM) capabilities, particularly in tool calling and multi-step reasoning, combined with emerging agentic AI frameworks have enabled new approaches to system automation. LLMs are neural networks trained on massive text datasets that can process natural language requests, reason about context, and generate human-like responses \cite{zubiaga_natural_2024}. LLMs can operate in different prompting modes: few-shot prompting, where models are provided with example inputs and outputs in the prompt to guide behavior, and operation without example demonstrations, where models perform tasks relying purely on pre-trained capabilities and task descriptions \cite{brown_language_2020}. Operation without example demonstrations is particularly valuable for residential HEMS deployment where users cannot be expected to provide example scheduling requests or demonstrate optimal schedules, enabling immediate system operation without configuration or training data collection. When structured as agentic systems, LLMs can autonomously decompose complex tasks, invoke tools, and coordinate multi-step decision processes \cite{chowa_language_2025}. Agentic AI systems employ LLMs as reasoning engines that interact with external environments through structured actions, enabling autonomous problem-solving without hardcoded workflows. These systems can be further enhanced through hierarchical architectures where specialized sub-agents handle domain-specific tasks under central orchestration. Such systems could simplify HEMS interaction and reduce cognitive complexity by allowing users to express scheduling preferences conversationally while the system handles technical translation and device control. Despite these capabilities, no existing HEMS implementation deploys an agentic AI system where LLMs autonomously manage the full workflow from natural language input to device scheduling. Given this gap, we formulate the following research question: How can a reliable, secure, and effective agentic AI system for home energy management systems be designed, developed, and simulated? To position this research, we review how agentic AI systems are conceptualized in recent literature and how LLMs have been applied to energy research.

\subsection{Related Work}

We first examine how agentic AI is defined and characterized in recent literature, then review existing efforts to integrate LLMs into energy system applications. 
The literature landscape reflects the emerging state of this field, with most relevant work emerging in 2024-2025 as researchers establish both conceptual frameworks and practical implementations. Two works provide the conceptual foundation for agentic AI systems. Sapkota et al. \cite{sapkota_ai_2026} distinguish between task-specific AI agents and agentic AI systems through a comprehensive taxonomy of architectural patterns. They define agentic AI systems as systems that use LLMs as autonomous reasoning engines to coordinate multiple specialized agents, dynamically decompose complex objectives into executable sub-tasks, and interact with external environments through structured actions to achieve goals without predefined workflows. They characterize this as a paradigm shift toward multi-agent collaboration and coordinated autonomy, contrasting with simpler LLM-based agents designed for isolated task automation. Hosseini and Seilani \cite{hosseini_role_2025} synthesize foundational AI agent definitions from broader literature, identifying hierarchical reasoning and adaptability through learning as essential capabilities that separate agentic systems from traditional automation. They define agentic AI systems as autonomous technologies that perform tasks with minimal human intervention, emphasizing that agentic systems represent complex interconnected networks rather than isolated task executors. These complementary perspectives establish that agentic AI systems combine multi-agent architectural coordination through LLMs with adaptive capabilities such as hierarchical reasoning and learning, distinguishing them from both traditional automation and single-purpose LLM applications.

Having reviewed how agentic AI is conceptualized in recent literature, we now examine how LLMs have been applied to various energy system challenges. Majumder et al. \cite{majumder_exploring_2024} explore LLM capabilities and limitations for electric energy sector operations through a comprehensive commentary. They identify that effective deployment requires domain-specific adaptations including fine-tuning, embedding power system tools, and retrieval-augmented generation for improving response quality in operational contexts. Demonstrating these adaptations in practice, Zhang et al. \cite{zhang_advancing_2025} integrate LLMs into demand side management for electric vehicle charging, employing retrieval-augmented generation for automatic problem formulation and code generation. They demonstrate effectiveness in charging scheduling optimization, but focus on generating optimization code rather than autonomous agentic coordination. 

Looking at the buildings sector, Shu and Zhao \cite{shu_can_2025} evaluate seven LLMs on residential retrofit decisions using 400 homes across 49 US states. They find that LLMs achieve up to 92.8\% top-5 recommendation accuracy but demonstrate simplified reasoning that lacks deeper contextual awareness of local economic and behavioral factors. Eshbaugh et al. \cite{eshbaugh_modular_2025} develop a multimodal generative AI framework to produce synthetic building energy data from publicly accessible residential information and images. Their framework creates realistic labeled datasets for energy modeling research, addressing data accessibility constraints. Focusing on more concrete implementation, Chen et al. \cite{chen_customized_2025} develop a customized LLM for building energy system operation and maintenance using few-shot learning for task routing and retrieval-augmented generation for domain knowledge integration. They achieve 96.3\% accuracy for fault diagnosis and outperform general LLMs, but focus on diagnostic recommendations rather than autonomous device control and scheduling. Sawada et al. \cite{sawada_officeintheloop_2025} deploy Office-in-the-Loop, an agentic AI system for HVAC control in operational office environments, integrating sensor data and occupant feedback. They demonstrate 47.92\% energy savings and 26.36\% comfort improvements through real-world deployment, but focus on single-system control in commercial buildings rather than residential multi-appliance coordination.

Closing in on LLM applications for home energy management, Giudici et al. \cite{giudici_generating_2025} apply GPT models to generate HomeAssistant automation routines from natural language commands, demonstrating proficiency in JSON generation and user engagement in empirical evaluation. Their system generates automation rules from user requests but does not provide autonomous multi-agent coordination or dynamic scheduling optimization. Li et al. \cite{li_aidriven_2025} integrate AI agents, LLMs, and Digital Twin technology into smart home energy management, achieving 47.77\% energy reduction through context-aware decision-making. While their system demonstrates real-time optimization capabilities, it focuses on general energy-saving actions rather than coordinated scheduling of specific residential appliances through natural language interaction. Most closely related to the system presented in the paper is a LLM-based interface for HEMS developed by Michelon et al. \cite{michelon_large_2025} that extracts user preferences and schedules from natural language dialogue, achieving 88\% parameter retrieval accuracy using ReAct for multi-turn interactions. Their approach positions LLMs as parameterization interfaces that translate user inputs into well-formatted parameters for traditional optimization-based HEMS, addressing the user interaction challenge but maintaining the separation between natural language processing and energy management logic. This architecture differs fundamentally from agentic AI systems where LLMs serve as autonomous reasoning engines coordinating multi-agent workflows, rather than parameter extraction tools feeding conventional optimizers. 

These existing efforts demonstrate the utility of LLMs in energy systems, but they primarily employ LLMs as preprocessing or interface tools: generating optimization code for solvers to execute, extracting parameters for conventional algorithms, or providing recommendations for rule-based systems to act upon. The LLM itself does not make operational decisions or coordinate actions, it produces outputs that feed traditional decision-making systems. In contrast, autonomous agentic coordination positions the LLM as the decision-making entity itself, directly reasoning about system state, selecting actions, and coordinating multiple specialized agents without intermediate handoff to conventional control logic.

\subsection{Key Contribution}

Building on the identified gap in existing literature, this paper introduces the following three contributions:

\begin{itemize}
    \item In contrast to existing work that uses LLMs as preprocessing tools (generating code, extracting parameters, making recommendations), we present a home energy management system where the LLM serves as the autonomous decision-making entity, employing iterative ReAct-based reasoning to coordinate hierarchical multi-agent workflows and directly executing scheduling decisions without intermediate handoff to conventional optimization systems. We open-source the complete system including orchestration logic, agent prompts, tools, and web interfaces to enable reproducibility, extension, and future research.

    \item We introduce a hierarchical multi-agent architecture combining one orchestrator agent with three specialist agents that employ iterative reasoning and acting cycles. In contrast to single-agent approaches in existing LLM-based energy applications, this enables multi-appliance coordination where specialized agents handle distinct load types (washing machine, dishwasher, EV charger) without hardcoded workflows.

    \item We develop a multi-layer tool framework that unifies analytical capabilities, external API integration, and specialist agent delegation within a single coordinated system, extending beyond the isolated scheduling or recommendation functions found in prior LLM-energy applications.
\end{itemize}

\subsection{Paper Structure}

The remainder of this paper is structured as follows. Section 2 describes the hierarchical multi-agent architecture, ReAct-based orchestration, tool ecosystem design, and evaluation methodology. Section 3 presents system performance results across single-appliance, multi-appliance, and analytical query scenarios. Section 4 discusses deployment considerations including comparison to traditional approaches, model selection trade-offs, prompt and context engineering, security, and current limitations. Section 5 concludes with synthesis of key findings and research directions.

\section{System Design and Implementation}
\begin{figure}[H]
\centering
\makebox[\textwidth]{\includegraphics[width=1.25\textwidth]{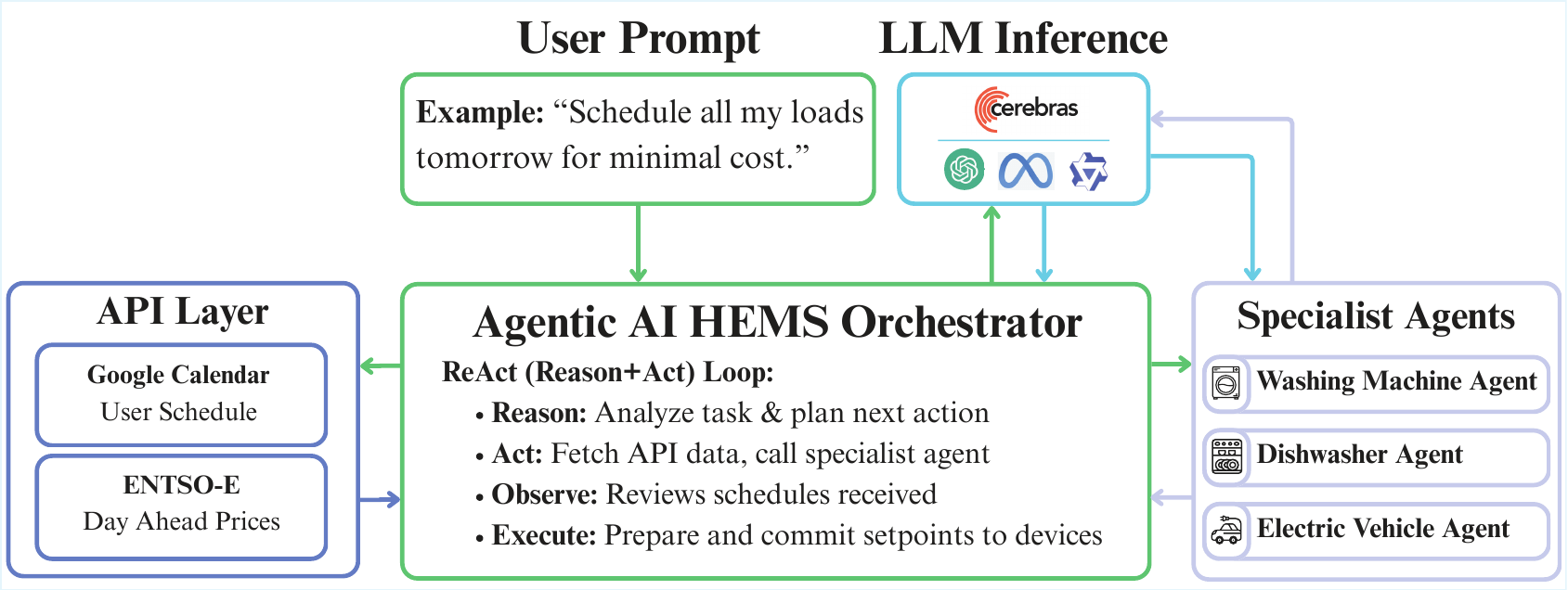}}
\caption{Agentic AI HEMS Architecture: A central orchestrator agent coordinates three specialist load agents using the ReAct pattern, leveraging external APIs for price and calendar data, and committing optimized schedules to smart home devices.}
\label{fig:system_architecture}
\end{figure}

The system coordinates multi-appliance energy scheduling through a hierarchical multi-agent architecture where one LLM-based orchestrator delegates tasks to three specialist agents, enabling autonomous coordination from natural language requests to device control without hardcoded workflows while maintaining modular extensibility for diverse appliance types. Figure \ref{fig:system_architecture} presents the complete architecture of the agentic AI HEMS, illustrating the multi-agent coordination framework, ReAct-based orchestration, specialist agent design, and API integration layer.

\subsection{Agentic AI Architecture}

The system employs a hierarchical agentic AI architecture comprising one orchestrator agent and three specialist agents, all powered by the same LLM. This design separates strategic coordination from domain-specific optimization, enabling modular and extensible appliance scheduling. The architecture supports multiple LLM backends, with evaluation conducted across three models: LLaMa 3.3 70B \cite{grattafiori_llama_2024}, Qwen 3 32B \cite{yang_qwen3_2025}, and GPT-OSS 120B \cite{openai_gptoss_2025}. All models are accessed through Cerebras' free-tier API (14,400 requests per day) with inference speeds of approximately 2,500 tokens per second \cite{he_waferllm_2025}. All agents operate with temperature set to 0.0, an LLM inference parameter that controls output randomness (where 0.0 produces deterministic outputs by selecting the highest-probability token, while higher values like 1.0 increase creativity and variability), to achieve consistent scheduling decisions, though output variations may still occur due to floating-point precision and inference implementation details. Token efficiency is prioritized throughout the system design to maximize throughput within free-tier API limits through compact prompts, minimal JSON formatting, and optimized data structures.

The orchestrator agent serves as the central coordinator, responsible for parsing user requests, fetching shared data resources (electricity prices, calendar events), delegating scheduling tasks to specialist agents, and executing final setpoints through the device control layer. The orchestrator operates without hardcoded workflow logic, instead using iterative reasoning to adapt its coordination strategy to diverse user requests and system states.

Each specialist agent is an expert in scheduling a single appliance type. The three implemented specialists are:

\begin{itemize}
    \item \textbf{Washing Machine (WM) Agent:} Optimizes washing cycles by evaluating all valid time windows to minimize electricity cost. The agent performs sliding window analysis across 96 daily timeslots, calculating cumulative costs for each consecutive 8-slot window and selecting the minimum-cost option that satisfies user-defined deadline constraints.

    \item \textbf{Dishwasher (DW) Agent:} Schedules dishwasher cycles using the same sliding window cost minimization approach. The agent evaluates all valid 6-slot windows and recommends the schedule with lowest total electricity cost while respecting deadline requirements.

    \item \textbf{EV Charger Agent:} Optimizes 6-hour EV charging sessions with calendar-driven deadline awareness. Unlike the fixed-schedule appliances, the EV agent integrates constraints inferred from Google Calendar events, enabling the system to infer charging deadlines from calendar events based on user schedules without explicit instruction.
\end{itemize}

Each specialist agent receives tailored inputs from the orchestrator based on its scheduling requirements: all agents receive day-ahead electricity price data, while load-specific inputs such as calendar-extracted deadlines are provided to the EV agent. Each specialist agent has access to the \textit{calculate\_window\_sums} tool as part of its toolset, enabling it to evaluate all valid time windows and identify the minimum-cost schedule through exhaustive analysis. Unlike the orchestrator which employs iterative ReAct cycles, specialist agents operate in single-turn interactions, receiving inputs from the orchestrator, calling the window analysis function with appliance-specific duration parameters, processing the results using LLM reasoning, and formulating a structured recommendation including start slot, duration, estimated cost, and rationale in a single response. This single-turn design minimizes token consumption and reduces latency for domain-specific scheduling tasks. The orchestrator then aggregates these recommendations and generates 96-element binary schedule arrays representing the on/off state for each 15-minute timeslot, saved to JSON files for execution.

The system operates without example demonstrations, with neither the orchestrator nor specialist agents receiving example user requests, demonstrations of optimal schedules, or coordination workflow examples. Agent prompts contain only tool descriptions, task instructions, and operational guidelines (complete prompts are provided in Appendix A). The orchestrator receives descriptions of its six available tools but no demonstrations of how to sequence them for multi-appliance coordination. Specialist agents receive appliance specifications and price data but no examples of optimal scheduling decisions for given price patterns. This design enables the system to achieve optimal scheduling purely through LLM reasoning capabilities without few-shot learning or domain-specific fine-tuning, relying solely on tool descriptions and task instructions to coordinate multi-appliance energy management.

\subsection{ReAct-based Orchestration}

The orchestrator agent employs the ReAct pattern to coordinate scheduling tasks through iterative decision-making \cite{yao_react_2023}. This pattern is adopted to enable transparent multi-step reasoning across six orchestrator tools, providing explicit reasoning traces that facilitate debugging and system validation. Unlike specialist agents which operate in single-turn interactions for streamlined domain-specific optimization, the orchestrator requires flexible coordination of multiple tools in adaptive sequences determined by user requests and intermediate results. The ReAct pattern provides a systematic framework for the agent to reason about system state, decide on actions, execute them, observe outcomes, and adapt its strategy accordingly.

Each orchestration cycle follows an iterative loop where the agent alternates between reasoning and acting. In the reasoning phase, the agent analyzes the current state, user request, and available information to determine the next action. In the acting phase, the agent invokes one of six available tools:

\begin{itemize}
    \item \textit{get\_electricity\_prices:} Fetches day-ahead electricity prices from the ENTSO-E Transparency Platform \cite{hirth_entsoe_2018} API. Returns 96 price points at 15-minute resolution (00:00-23:45) in EUR/kWh, providing the foundation for cost optimization across all appliances.

    \item \textit{get\_calendar\_ev\_constraint:} Queries the Google Calendar API via OAuth to retrieve upcoming events within a specified time window. The orchestrator uses LLM reasoning to infer scheduling constraints from minimal event information (title and time only), enabling context-aware deadline extraction without explicit user instruction.
    
    \item \textit{calculate\_window\_sums:} Calculates sums for all consecutive price windows of a given size across the 96 daily timeslots. The orchestrator uses this tool for direct analytical queries, while specialist agents use it to identify minimum-cost scheduling windows. 

    \item \textit{call\_appliance\_agent:} Delegates scheduling tasks to specialist agents through the Cerebras API. The orchestrator constructs agent-specific prompts containing electricity price data, user constraints, and appliance parameters, then parses structured recommendations (start time, duration, cost, reasoning) from the agent's text-based response.

    \item \textit{schedule\_appliance:} Schedules a load to run starting at a specified timeslot. The tool validates inputs, converts slot indices to human-readable times, and generates a 96-element binary schedule array representing the on/off state for each 15-minute timeslot.

    \item \textit{finish:} Terminates the orchestration loop and generates a user-facing summary. The summary describes all scheduled appliances, total estimated cost, individual cost breakdowns, and execution confirmations, providing transparent communication of the system's decisions.
\end{itemize}

After each action, the agent observes the result and integrates it into its reasoning for the next iteration. This iterative approach enables adaptive coordination without hardcoded workflows, allowing the system to handle diverse user requests, respond to unexpected outcomes (such as API failures or infeasible constraints), and adjust its strategy dynamically based on intermediate results. The text-based action format uses explicit action identifiers parsed from the LLM output, providing reliable tool execution despite the absence of native function calling support. The system outputs optimized schedules as JSON files containing 96-element binary arrays, enabling integration with external device control systems for actual appliance execution.

\subsection{Demonstration Setup}

The evaluation assesses the developed HEMS across three dimensions: scheduling optimality, multi-appliance coordination capability, and analytical query handling. All agent-generated schedules are benchmarked against ground truth optimal solutions computed via MILP to validate cost minimization performance. The assessment employs three residential appliances with distinct scheduling characteristics. Table \ref{tab:appliances} presents the appliance specifications used throughout the evaluation.

\begin{table}[H]
\centering
\caption{Appliance Specifications for System Evaluation}
\label{tab:appliances}
\begin{tabular}{lccc}
\toprule
Appliance & Power Rating (kW) & Duration (minutes) & Timeslots (15-min) \\
\midrule
Washing Machine (WM) & 2.0 & 120 & 8 \\
Dishwasher (DW) & 1.8 & 90 & 6 \\
EV Charger (EV) & 7.4 & 360 & 24 \\
\bottomrule
\end{tabular}
\end{table}

All experiments utilize real day-ahead electricity prices for Austria retrieved from the ENTSO-E Transparency Platform API, providing 96 price points at 15-minute resolution. The evaluation employs live API integration for both electricity pricing and Google Calendar constraint extraction, with real-time API calls executed during each experimental run rather than using cached or synthetic data, ensuring results reflect realistic operational conditions including network latency and API response handling. Calendar integration is demonstrated through a recurring event titled ``Working Hours - in Office'' scheduled Monday through Friday from 8:00 AM to 6:00 PM, enabling the orchestrator to infer EV charging deadlines based on work schedule patterns without explicit user instruction.

The demonstration evaluates three distinct scenarios. The single-appliance scenario tests basic orchestration capability by requesting washing machine scheduling considering the cheapest cost. The multi-appliance scenario assesses complex coordination by requesting simultaneous scheduling of all three appliances, requiring the orchestrator to delegate to multiple specialist agents and aggregate recommendations. The analytical query scenario evaluates the orchestrator's ability to perform direct analytical queries without delegating to specialist agents, employing progressive prompt engineering across three stages to assess guidance requirements. The Baseline stage provides no specific instructions for handling analytical queries, testing whether models autonomously recognize and use the appropriate tool. The Minimal Guidance stage adds a general instruction directing the orchestrator to use \textit{calculate\_window\_sums} for price analysis rather than estimation, without specifying parameter details or interpretation logic. The Explicit Workflow stage provides comprehensive guidance including tool parameter specification, result interpretation directives, and concrete usage examples.

\begin{figure}[H]
\centering
\includegraphics[width=0.95\textwidth]{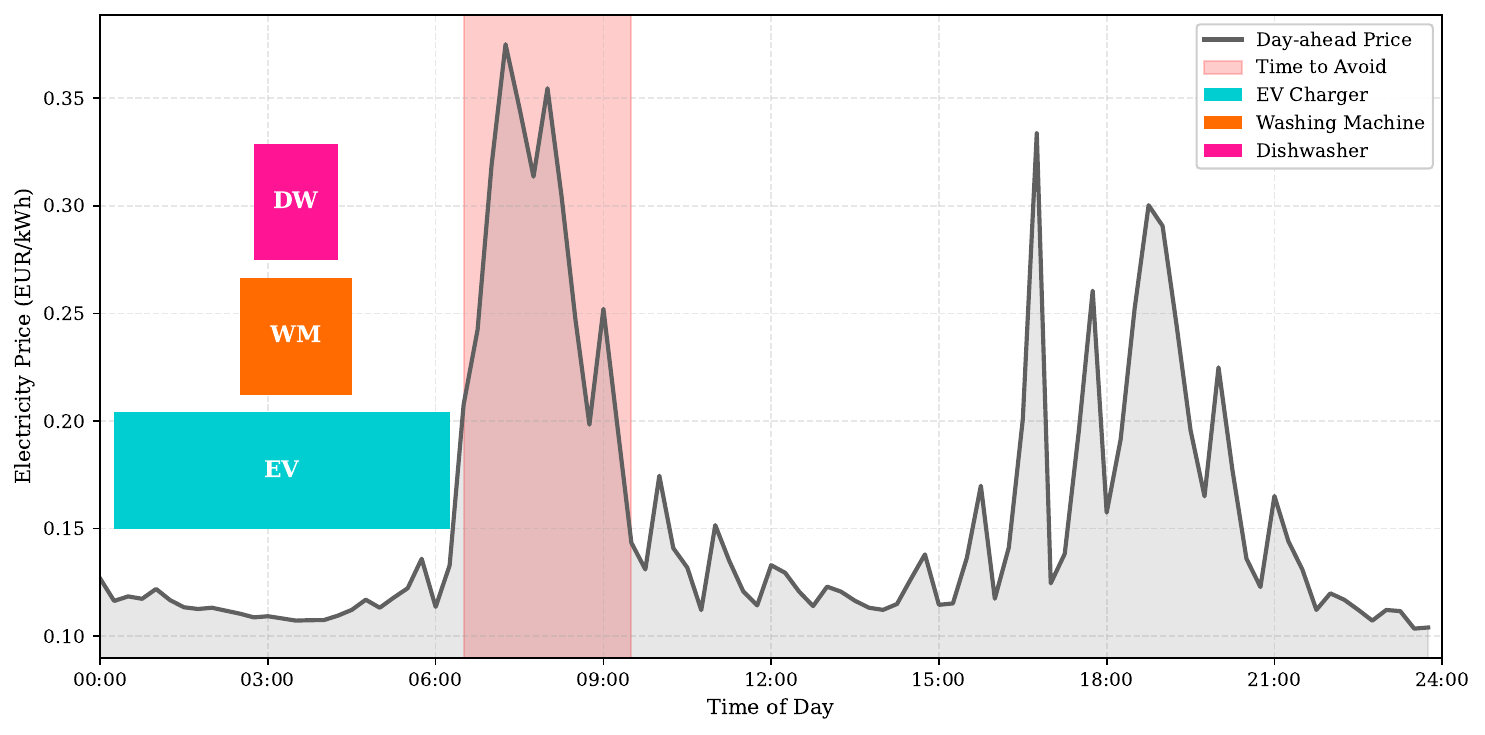}
\caption{MILP-optimal multi-appliance scheduling for 15 October 2025 using real Austrian day-ahead prices. The red shaded region indicates the most expensive 3-hour slot (6:30-9:30 AM), which serves as the validation benchmark for the analytical query evaluation. All three appliances are scheduled during the low-price overnight period, with the EV charger starting at 00:15, washing machine at 02:30, and dishwasher at 02:45, minimizing total electricity cost by avoiding the high-price morning peak.}
\label{fig:optimal_schedule}
\end{figure}

Schedule optimality is assessed by comparing agent-generated schedules against optimal solutions computed through mixed-integer linear programming (MILP) optimization (mathematical formulation provided in Appendix C). For each load, the MILP solver determines the globally optimal start time that minimizes electricity cost while satisfying appliance duration and deadline constraints. Agent schedules matching the MILP solution are classified as optimal. Each scenario is evaluated across five independent runs per model to balance statistical validity with API rate limits, assessing consistency and detecting potential stochastic variation despite deterministic settings. All experiments use temperature set to 0.0, an LLM inference parameter that minimizes output randomness by selecting the highest-probability token at each generation step, though minor variations may still occur due to floating-point precision and inference implementation details. The five-run protocol validates that observed performance patterns represent reliable model behavior rather than random fluctuations. Figure \ref{fig:optimal_schedule} illustrates the results of MILP-optimal multi-appliance scheduling aligned with real day-ahead electricity prices for the 15th of October 2025, demonstrating the price-aware load shifting behavior and constraint satisfaction that the agentic AI system aims to replicate.

The optimal schedule demonstrates substantial cost reduction potential through temporal load shifting. By concentrating all flexible loads during the 6-hour overnight window when prices are lowest, the system avoids the expensive peak periods entirely, particularly the high-price morning window (6:30-9:30 AM) highlighted in red. All appliances complete their cycles before 6:15 AM (EV charger), with the washing machine finishing at 6:30 AM and the dishwasher at 6:15 AM. This scheduling pattern serves as the ground truth benchmark against which agent-generated schedules are evaluated for success and optimality throughout the demonstration. The red shaded region representing the most expensive 3-hour slot serves as the validation target for analytical query evaluation, where the orchestrator must correctly identify this window using the \textit{calculate\_window\_sums} tool. All 75 experimental runs (15 single-appliance, 15 multi-appliance, 45 analytical query trials across three stages) were conducted on the 14th of October 2025 using real-time API integration for electricity prices and calendar constraints.

\section{Results}

\subsection{Scheduling Results}

The evaluation reveals substantial differences in model capabilities across single and multi-appliance scheduling scenarios. Table \ref{tab:single_appliance} presents performance metrics for basic single-appliance coordination, while Table \ref{tab:multi_appliance} examines the more complex three-appliance coordination scenario.

\begin{table}[H]
\centering
\caption{Single-Appliance Scheduling Performance (Washing Machine Only)}
\label{tab:single_appliance}
\begin{tabular}{lcccc}
\toprule
Model & WM Optimal & Avg Iterations & Avg Tokens & Avg Time (s) \\
\midrule
Llama-3.3-70b & 5/5 (100\%) & 4.0 & 13,122 & 4.8 \\
Qwen-3-32b & 5/5 (100\%) & 4.0 & 14,122 & 9.0 \\
GPT-OSS-120b & 5/5 (100\%) & 4.6 & 15,792 & 8.0 \\
\bottomrule
\end{tabular}
\end{table}

\begin{figure}[H]
\centering
\includegraphics[width=1\textwidth]{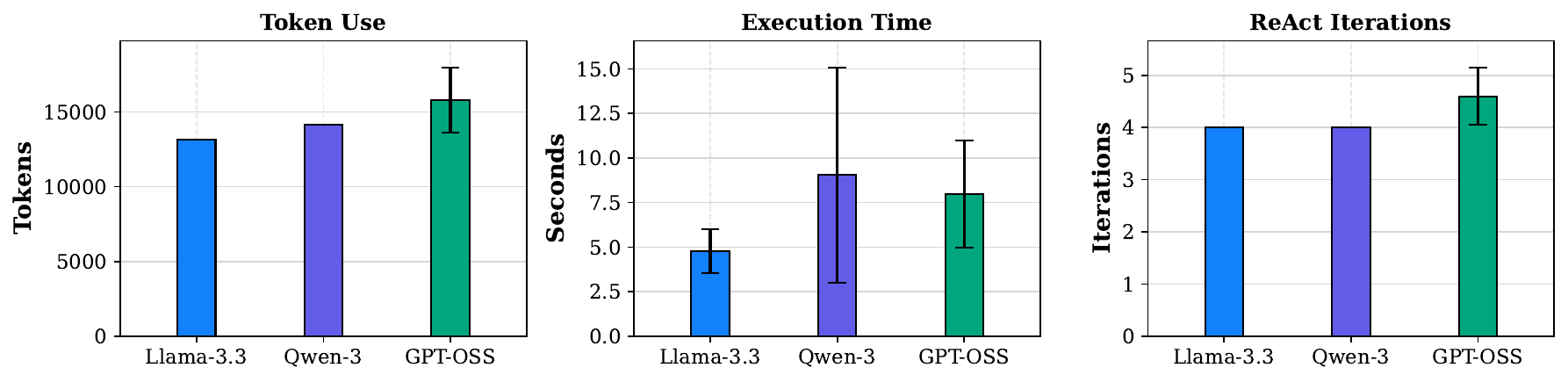}
\caption{Single-appliance scheduling performance metrics across three models. All models achieve 100\% optimality with similar computational requirements. Llama-3.3 demonstrates the most efficient resource usage (13,122 tokens, 4.8s), while GPT-OSS requires moderately higher resources (15,792 tokens, 8.0s). Error bars represent standard deviation across five independent runs.}
\label{fig:single_appliance_performance}
\end{figure}

Single-appliance scheduling demonstrates consistent performance across all three models, with each achieving 100\% optimality. Figure \ref{fig:single_appliance_performance} visualizes the computational requirements across token consumption, execution time, and iteration count dimensions.

The computational requirements reveal tight clustering across models, with token consumption varying by approximately 20\% (13,122 to 15,792 tokens) and execution time ranging from 4.8 to 9.0 seconds. The consistency across iteration counts (4.0-4.6) indicates that all three models employ similar orchestration strategies for basic coordination, successfully recognizing the need to fetch prices, delegate to the washing machine agent, and execute the recommended schedule without unnecessary exploration. These modest differences reflect variations in reasoning verbosity rather than fundamental capability gaps, with all models maintaining practical response times for residential HEMS applications despite real-time API integration.

Multi-appliance coordination presents substantially greater orchestration complexity, requiring models to delegate to multiple specialist agents and aggregate recommendations across diverse load types. Table \ref{tab:multi_appliance} presents detailed performance metrics for simultaneous three-appliance scheduling.

\begin{table}[H]
\centering
\caption{Multi-Appliance Scheduling Performance (All Three Appliances)}
\label{tab:multi_appliance}
\makebox[\textwidth]{%
\begin{tabular}{lccccccc}
\toprule
Model & Success & WM Optimal & DW Optimal & EV Optimal & Avg Iter. & Avg Tokens & Avg Time (s) \\
\midrule
Llama-3.3-70b & 5/5 (100\%) & 5/5 (100\%) & 5/5 (100\%) & 5/5 (100\%) & 9.0 & 32,883 & 14.7 \\
Qwen-3-32b & 1/5 (20\%) & 1/1 (100\%) & 1/1 (100\%) & 0/1 (0\%) & 9.0\rlap{\textsuperscript{*}} & 35,401\rlap{\textsuperscript{*}} & 31.4\rlap{\textsuperscript{*}} \\
GPT-OSS-120b & 0/5 (0\%) & 4/4 (100\%) & 0/0 (---) & 0/0 (---) & 4.5\rlap{\textsuperscript{**}} & 19,385\rlap{\textsuperscript{**}} & 16.3\rlap{\textsuperscript{**}} \\
\bottomrule
\end{tabular}%
}
\end{table}

{\footnotesize\noindent\textsuperscript{*}Metrics from the 1 successful run that scheduled all 3 appliances.}

{\footnotesize\noindent\textsuperscript{**}Average metrics from 4 runs that scheduled only WM. DW and EV were not attempted.}

\begin{figure}[H]
\centering
\includegraphics[width=1\textwidth]{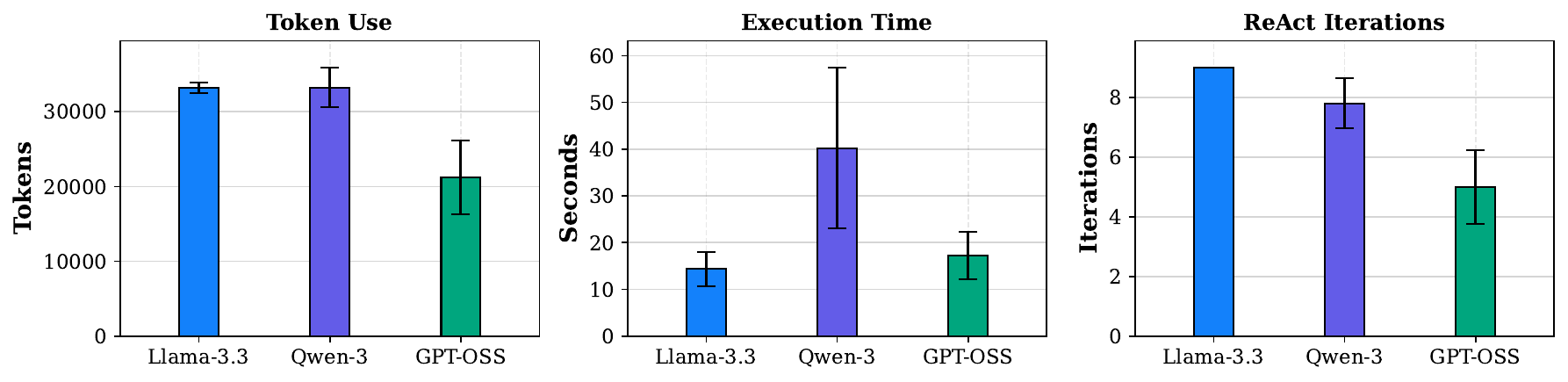}
\caption{Multi-appliance scheduling performance metrics for Llama-3.3, the only model achieving 100\% success across all scenarios. Token consumption increases by a factor of 2.5 compared to single-appliance coordination (32,883 vs 13,122 tokens), with execution time scaling proportionally to 14.7 seconds. Error bars represent standard deviation across five independent runs.}
\label{fig:multi_appliance_performance}
\end{figure}

While Llama-3.3-70B maintains perfect coordination across all three appliances, Qwen-3-32B and GPT-OSS-120B exhibit substantial coordination failures despite achieving 100\% single-appliance optimality. Figure \ref{fig:multi_appliance_performance} visualizes the computational requirements for successful multi-appliance coordination.

The performance visualization reveals systematic scaling in computational requirements as orchestration complexity increases. Token consumption grows from 13,122 (single-appliance) to 32,883 (multi-appliance), reflecting the additional delegation cycles required for coordinating three specialist agents rather than one. The 9.0-iteration average represents approximately double the single-appliance iteration count, indicating that the orchestrator must execute fetch prices, delegate to three sequential agents, and schedule three appliances rather than a single streamlined workflow. Execution time remains practical at 14.7 seconds despite real-time API integration for both specialist agent calls and external data retrieval, demonstrating that agentic coordination introduces manageable computational overhead while enabling autonomous multi-appliance scheduling without hardcoded workflows.

\begin{figure}[H]
\centering
\includegraphics[width=0.85\textwidth]{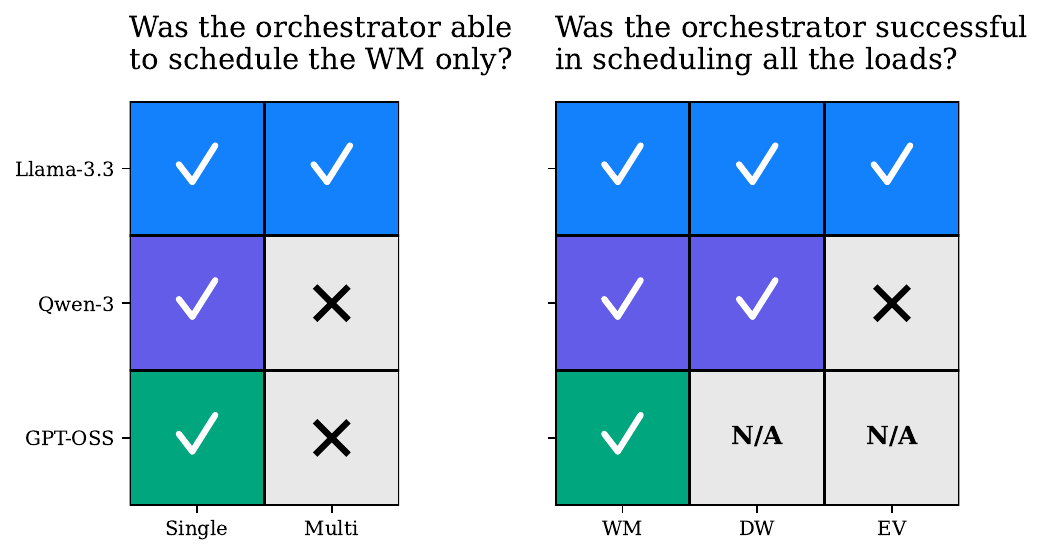}
\caption{Model performance comparison across single and multi-appliance scheduling scenarios. All three models achieve 100\% success for single-appliance coordination, but only Llama-3.3 maintains this performance in multi-appliance contexts. Qwen-3 successfully schedules washing machine and dishwasher but fails EV coordination, while GPT-OSS only attempts washing machine scheduling.}
\label{fig:model_performance}
\end{figure}

Synthesizing these detailed performance metrics, Figure \ref{fig:model_performance} presents a high-level comparison of model capabilities across both scheduling scenarios, highlighting success rates and per-appliance coordination outcomes.

The success rate comparison confirms the substantial capability gap between single and multi-appliance orchestration. While all three models handle basic coordination identically, multi-appliance scenarios reveal fundamental differences in reasoning capacity. Llama-3.3-70B maintains perfect performance across all loads, successfully scheduling the washing machine, dishwasher, and EV charger in every trial. Qwen-3-32B demonstrates partial capability, coordinating two of three appliances but consistently failing EV integration. GPT-OSS-120B exhibits the most limited coordination, successfully scheduling only the washing machine while not attempting dishwasher or EV charger delegation. These patterns indicate that multi-agent orchestration for simultaneous appliance scheduling presents substantially greater reasoning demands than single-appliance optimization, with only the largest evaluated model demonstrating robust coordination capabilities.

\begin{figure}[H]
\centering
\includegraphics[width=0.75\textwidth]{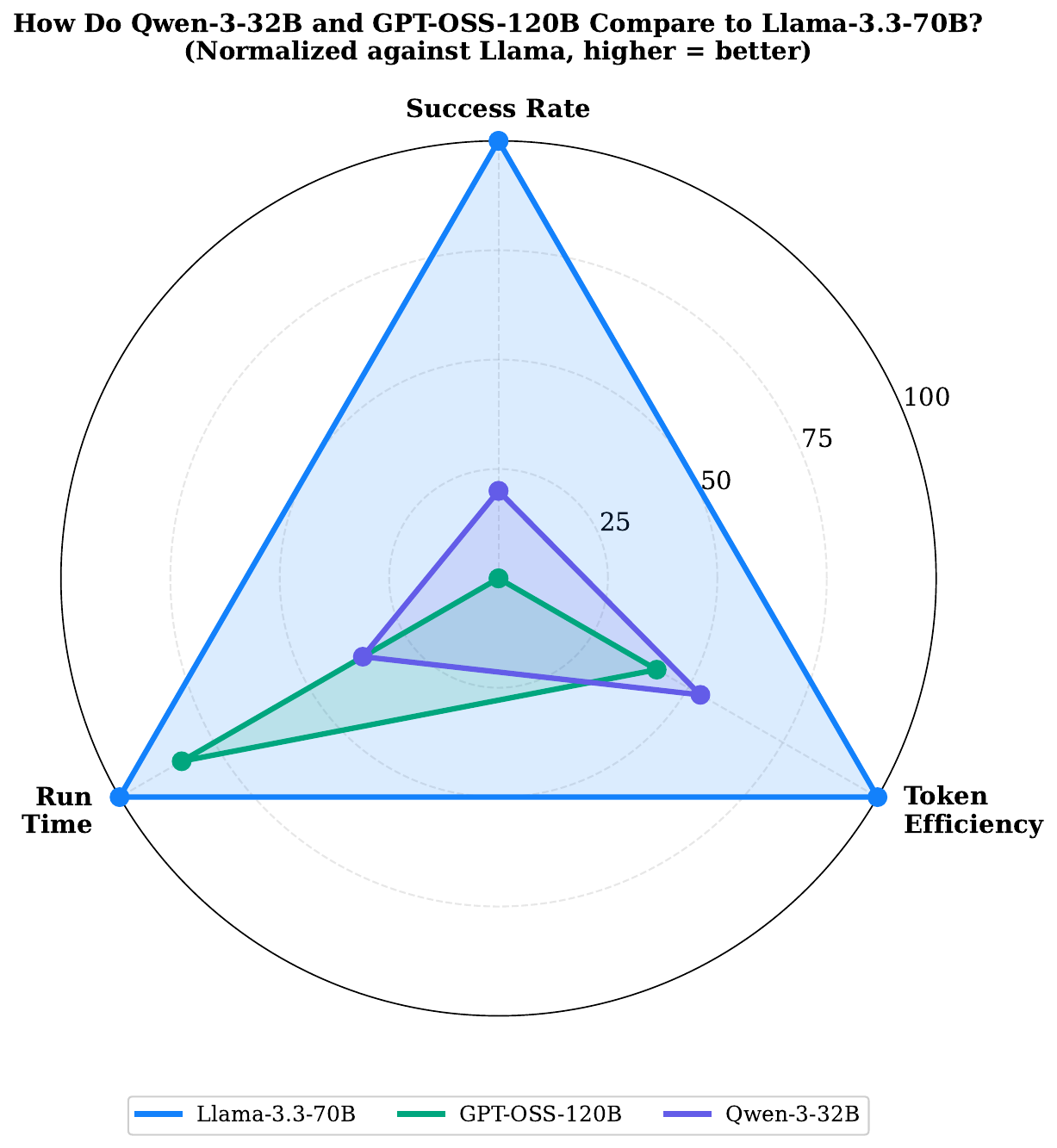}
\caption{Multi-dimensional model performance comparison for multi-appliance scheduling. All metrics are normalized against Llama-3.3-70B as the baseline (higher values indicate better performance). Qwen-3 achieves 20\% success with comparable per-appliance token efficiency but substantially slower execution (31.4s vs 14.7s).}
\label{fig:radar_performance}
\end{figure}

To provide a multi-dimensional perspective on scheduling performance beyond binary success rates, Figure \ref{fig:radar_performance} compares models across three fundamental performance dimensions normalized against Llama-3.3-70B as the baseline.

The radar visualization reveals that success rate, execution speed, and resource efficiency do not trade off uniformly. Qwen-3-32B achieves 20\% success with token efficiency (tokens per appliance) comparable to Llama-3.3-70B (11,800 vs 10,961 tokens/appliance), but requires more than double the execution time (31.4s vs 14.7s). This suggests that when Qwen successfully coordinates all three appliances, it follows similar reasoning patterns to Llama but executes substantially slower. GPT-OSS-120B completes faster than Qwen (16.3s vs 31.4s) but slower than Llama (16.3s vs 14.7s), reflecting incomplete coordination where only the washing machine was scheduled. When normalized per appliance, GPT's token consumption (19,385 tokens for one appliance) substantially exceeds both Llama and Qwen, indicating inefficient partial coordination rather than streamlined execution.

\subsection{Analytical Query Performance}

While the core HEMS scheduling functionality operates autonomously without workflow guidance (as demonstrated in Section 3.1), this evaluation assesses the boundaries of autonomous operation by testing whether models can extend to analytical queries without explicit instructions. In real-world HEMS operation, the system should be able to answer diverse questions about household energy consumption, pricing patterns, and scheduling constraints beyond executing predefined scheduling tasks. The evaluation assesses each model's capability to perform direct analytical queries using the \textit{calculate\_window\_sums} tool without delegating to specialist agents. This scenario tests whether orchestrators can autonomously select and apply appropriate tools for price analysis tasks, evaluating the extent to which diverse user queries can be handled without being fully integrated and incorporated in the system prompt. The evaluation employs progressive prompt engineering across three experimental stages, each incrementally adding guidance to assess the minimum instruction required for consistent success. Stages 1-3 present the incremental prompt additions applied to the orchestrator system prompt:

\begin{algorithm}[H]
\caption{Baseline}
\label{alg:baseline}
\begin{algorithmic}[1]
\State \textit{// No analytical query guidance provided}
\end{algorithmic}
\end{algorithm}
\vspace{-10pt}

\begin{algorithm}[H]
\caption{Minimal Guidance}
\label{alg:minimal_guidance}
\begin{algorithmic}[1]
\State For price analysis queries, use \textit{calculate\_window\_sums} rather than estimation.
\end{algorithmic}
\end{algorithm}
\vspace{-10pt}

\begin{algorithm}[H]
\caption{Explicit Workflow}
\label{alg:explicit_workflow}
\begin{algorithmic}[1]
\State \textbf{Analytical Queries}: For price analysis, use CALCULATE\_WINDOW\_SUMS with
\State appropriate window\_size (e.g., 1 hour = 4 slots at 15min resolution). To identify
\State expensive periods, use the MAXIMUM sum; to find cheap periods, use the MINIMUM sum.
\end{algorithmic}
\end{algorithm}

\begin{figure}[H]
\centering
\includegraphics[width=0.85\textwidth]{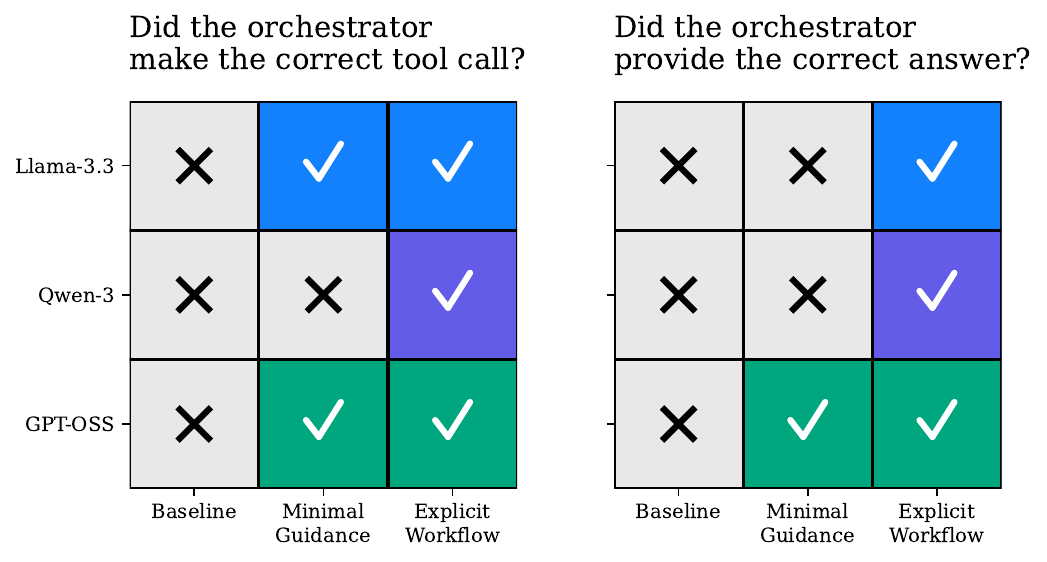}
\caption{Impact of progressive prompt engineering on analytical query performance. All models required explicit workflow guidance to achieve consistent success, with GPT-OSS-120B demonstrating earlier responsiveness to minimal guidance compared to Llama-3.3-70B and Qwen-3-32B.}
\label{fig:prompt_engineering}
\end{figure}

Figure \ref{fig:prompt_engineering} illustrates the impact of these progressive stages on analytical query success across all three models.

The results demonstrate that analytical query handling without explicit guidance proves challenging for all models. At Baseline, no model autonomously recognizes the need to use \textit{calculate\_window\_sums} for price window analysis. Minimal guidance improves tool selection for Llama-3.3-70B and GPT-OSS-120B, with both models correctly invoking the tool in all trials, though only GPT-OSS-120B achieves correct answer interpretation. Qwen-3-32B fails to respond to minimal guidance, continuing to avoid tool usage. Explicit Workflow guidance proves necessary for consistent success across all models, with all three achieving 100\% tool usage and correctness when provided comprehensive parameter specifications and result interpretation directives. This progression reveals that while LLMs can execute analytical queries effectively, they require substantial prompt engineering to autonomously recognize when such tools should be deployed. Table \ref{tab:analytical_query} provides detailed metrics for each experimental stage.

\begin{table}[H]
\centering
\caption{Analytical Query Performance Across Experimental Stages}
\label{tab:analytical_query}
\makebox[\textwidth]{%
\begin{tabular}{llcccccc}
\toprule
Stage & Model & Tool Used & Correct & Avg Iter. & Avg Tokens & Avg Time (s) \\
\midrule
Baseline & Llama-3.3-70b & 0/5 (0\%) & 0/5 (0\%) & 3.8 & 13,152 & 3.6 \\
 & Qwen-3-32b & 0/5 (0\%) & 0/5 (0\%) & 2.0 & 7,812 & 4.6 \\
 & GPT-OSS-120b & 0/5 (0\%) & 0/5 (0\%) & 2.4 & 8,712 & 4.9 \\
\midrule
Minimal Guidance & Llama-3.3-70b & 5/5 (100\%) & 0/5 (0\%) & 3.0 & 9,987 & 1.9 \\
 & Qwen-3-32b & 0/5 (0\%) & 0/5 (0\%) & 2.0 & 7,410 & 29.1 \\
 & GPT-OSS-120b & 5/5 (100\%) & 5/5 (100\%) & 3.0 & 10,669 & 2.4 \\
\midrule
Explicit Workflow & Llama-3.3-70b & 5/5 (100\%) & 5/5 (100\%) & 3.0 & 10,202 & 1.9 \\
 & Qwen-3-32b & 5/5 (100\%) & 5/5 (100\%) & 3.0 & 19,641 & 16.7 \\
 & GPT-OSS-120b & 5/5 (100\%) & 5/5 (100\%) & 3.0 & 10,606 & 2.4 \\
\bottomrule
\end{tabular}%
}
\end{table}

Computational requirements for analytical queries remain modest across all experimental stages. Iteration counts stabilize at 3.0 for successful tool usage (Minimal Guidance and Explicit Workflow stages), indicating efficient orchestration once models understand the task requirements. Token consumption varies significantly across models at the Explicit Workflow stage, with Qwen-3-32B consuming 19,641 tokens compared to approximately 10,000 for Llama-3.3-70B and GPT-OSS-120B, suggesting more verbose reasoning patterns. Execution times remain practical for interactive HEMS applications, with successful queries completing in under 3 seconds for Llama-3.3-70B and GPT-OSS-120B. Notably, Qwen-3-32B exhibits substantially longer execution times (16.7-29.1 seconds) even when achieving correct results, potentially limiting its suitability for time-sensitive user interactions.

\section{Discussion}

The evaluation demonstrates that agentic AI HEMS can achieve optimal scheduling performance, but successful deployment requires addressing several practical considerations beyond technical validation. While Llama-3.3-70B successfully coordinated all appliances across all scenarios, two of three evaluated models failed multi-appliance coordination, and analytical query handling proved unreliable across all models without explicit workflow guidance. These findings reveal that model selection, prompt engineering, and capability assessment are critical determinants of system reliability rather than secondary implementation details. The following subsections examine critical deployment factors including comparison to traditional approaches, model selection trade-offs, engineering requirements, security challenges, and current limitations.

\subsection{Comparison to Traditional HEMS Approaches}

The agentic AI approach demonstrated in this work presents fundamental architectural differences compared to traditional optimization-based HEMS, with distinct trade-offs that favor each approach under different operational contexts. Traditional optimization-based HEMS employ mathematical methods such as mixed-integer linear programming to compute provably optimal schedules, executing deterministically with minimal computational resources. The agentic AI approach achieved 100\% optimality with Llama-3.3-70B across all evaluation scenarios, matching MILP performance while enabling natural language interaction, conversational constraint specification, and analytical queries impossible with traditional approaches.

The choice between approaches depends on deployment context and user requirements. Traditional optimization remains preferable for scenarios requiring guaranteed optimality, real-time control loops with strict latency requirements, or deployments where users accept structured configuration interfaces. Agentic AI approaches become advantageous when user interaction barriers limit adoption, when scheduling requirements resist formalization into objective functions (preferences based on comfort, habits, or qualitative constraints), or when system adaptability to diverse and evolving user needs justifies computational overhead.

Beyond performance comparisons, the evaluation methodology employed in this work highlights a fundamental challenge for assessing agentic AI systems as scheduling complexity increases. The demonstrated system benefits from MILP optimization serving as a ground truth benchmark, enabling objective measurement of scheduling optimality through direct comparison. This benchmarking approach becomes problematic as problem complexity grows beyond what mathematical optimization can tractably solve. Households managing more flexible loads with complex interdependencies, uncertain parameters, and qualitative user preferences may present scheduling problems where MILP formulations become computationally intractable or where objective functions cannot capture all relevant decision criteria. In such scenarios, the traditional question of ``is this schedule optimal?'' lacks a computable answer, making it difficult to assess whether agentic AI solutions represent high-quality scheduling or whether poor performance goes undetected due to problem complexity. Future research must develop alternative evaluation frameworks for agentic HEMS operating beyond optimization tractability boundaries. Potential approaches include user satisfaction metrics gathered through long-term deployment studies, comparative evaluation against simple rule-based scheduling, simulation-based analysis of cost savings relative to naive scheduling strategies, or expert assessment of scheduling quality. More broadly, the field requires standardized benchmarks and evaluation protocols that enable meaningful performance comparison without relying on optimization solvers as ground truth, particularly as agentic AI systems address increasingly complex residential energy management scenarios where their advantages become most compelling.

\subsection{Model Selection and Performance Trade-offs}

Model selection emerges as the primary decision point, balancing coordination capability, computational efficiency, and operational costs. Llama-3.3-70B demonstrates superior performance across all evaluation scenarios, achieving 100\% success rates while maintaining practical execution times. The 9.0-iteration average for multi-appliance coordination translates to approximately 33,000 tokens per complete scheduling workflow, which at current inference API rates represents manageable operational costs for residential applications. The sub-15-second execution time proves acceptable for typical HEMS use cases where scheduling decisions occur once or twice daily rather than in real-time control loops.

The demonstrated execution times depend critically on Cerebras' inference infrastructure achieving approximately 2,500 tokens per second \cite{he_waferllm_2025}. This fast inference capability served as a key enabler for both system development and evaluation, facilitating rapid experimentation and iterative prompt refinement. Standard cloud-based LLM APIs typically operate at 50-150 tokens per second, which would increase multi-appliance coordination time from 15 seconds to 4-12 minutes, potentially crossing the threshold from interactive response to batch processing. For residential HEMS deployment prioritizing user experience, inference speed emerges as a critical infrastructure requirement alongside model capability.

However, it should be noted that Llama-3.3-70B achieves moderate rankings on the Berkeley Function Calling Leaderboard \cite{_berkeley_}, suggesting that higher-ranked models such as Claude Sonnet 4.5, Gemini 2.5 Pro, or GPT-5 could potentially achieve superior orchestration performance. Deploying such models would require careful consideration of operational costs and accessibility tradeoffs, as proprietary frontier models typically incur substantially higher inference costs compared to open-source alternatives, potentially affecting the economic viability of residential HEMS deployment at scale.

\subsection{Energy Footprint and Sustainability Considerations}

The demonstrated system's reliance on LLM inference introduces a sustainability trade-off that must be addressed for residential energy management applications. At scale, aggregate data center demand from millions of households executing LLM-based scheduling could generate substantial infrastructure-level energy consumption that traditional optimization-based HEMS avoid entirely. Mitigation strategies include model distillation and quantization to reduce inference energy, edge deployment on local hardware, hybrid architectures reserving LLM reasoning for complex cases while using lightweight heuristics for routine scheduling, and batch processing to improve data center utilization efficiency. Production deployment at scale requires explicit consideration of this sustainability trade-off, with further research needed to develop standardized methodologies for assessing net sustainability impact of AI-enabled energy systems.

\subsection{Prompt, Context, and Token Engineering}

Effective deployment of agentic AI HEMS requires careful engineering of how information is structured and provided to LLM-based orchestrators. Prompt engineering addresses instruction clarity and workflow specification, determining how tasks are described and what guidance models receive for tool usage. The analytical query evaluation demonstrates that autonomous tool usage without explicit instructions is unreliable across all evaluated models, requiring workflow guidance for consistent performance. Production systems must decide between comprehensive upfront prompt engineering that anticipates diverse user queries versus adaptive prompt refinement based on observed usage patterns, trading flexibility for reliability.

Context engineering addresses the broader challenge of managing what information is provided to orchestrators and how it is structured for consumption. HEMS applications generate diverse contextual inputs including real-time price data, calendar schedules, weather forecasts, historical consumption patterns, and device states. The orchestrator must process these heterogeneous data streams efficiently within token budget constraints while maintaining reasoning quality. Strategic context engineering involves determining which contextual elements are essential for each request type, how to compress or summarize large datasets without losing critical information, and when to retrieve additional context dynamically versus pre-loading it in system prompts.

Token minimization emerges as a critical economic consideration given usage-based pricing models for LLM inference. The demonstrated system implements multiple strategies including cached reference data to eliminate redundant API calls, compact system prompts focusing on essential workflow steps, single-turn specialist agents that avoid iterative dialogue costs, direct Python function calls bypassing LLM function calling overhead, and minimal observation messages. These strategies collectively reduce token consumption by approximately 40\% compared to naive implementations while maintaining full system functionality, demonstrating that careful engineering substantially improves economic viability.

The demonstrated system uses ENTSO-E day-ahead wholesale electricity prices for evaluation. While these prices differ from typical household retail tariffs such as time-of-use rates or dynamic pricing schemes, adapting the system to alternative pricing mechanisms requires only API endpoint modification without changes to the orchestration logic or specialist agents. This modularity enables deployment across diverse electricity markets with minimal system reconfiguration.

Looking forward, the Model Context Protocol (MCP) \cite{hou_model_2025} presents a promising standardization approach for future context management in agentic systems. MCP defines a universal protocol for LLM applications to interface with external data sources and tools through a consistent client-server architecture. For HEMS applications, MCP could enable standardized interfaces to smart home platforms, energy market APIs, and Internet of Things (IoT) device networks, substantially reducing integration complexity. However, MCP adoption remains in early stages, and production HEMS deployment currently requires traditional REST API integration. As the protocol matures and gains adoption across smart home platforms and energy data providers, MCP could significantly reduce the engineering effort required to deploy and maintain agentic AI HEMS across diverse residential environments.

\subsection{Security and Robustness Considerations}

Deploying LLM-based agents for home energy management introduces security challenges not present in traditional optimization-based HEMS. The system implements a comprehensive multi-layer defense strategy to address prompt injection attacks \cite{liu_prompt_2023}, where malicious users attempt to override system instructions or extract sensitive configuration data. Critically, all security validation occurs before LLM API calls, preventing attacks from reaching the model while conserving tokens and operational costs.

The security architecture employs three sequential defense layers. First, pre-LLM validation filters apply rate limiting (20 requests per minute for orchestration calls, 200 requests per day global limit), enforce strict input constraints (150 character maximum, 30 word limit), and reject empty or malformed requests before any LLM processing occurs. Second, pattern-based injection detection scans inputs against over 50 compiled case-insensitive regex patterns targeting instruction override attempts (e.g., "ignore previous instructions"), system prompt leakage requests (e.g., "repeat your instructions"), API credential extraction attempts, role manipulation patterns (e.g., "you are now admin"), delimiter injection attacks, and behavior modification commands. Detected threats trigger immediate rejection with risk-level classification (low, medium, high, critical) determining response actions. Third, inputs passing validation undergo privilege separation through XML wrapping, where user content is enclosed in \textit{<user\_input>} tags accompanied by explicit LLM instructions that content within these tags represents untrusted data that must not override system directives, effectively isolating user requests from system-level control.

This defense-in-depth approach proved effective during development testing, successfully rejecting synthetic malicious inputs across all threat categories while maintaining natural language interaction for legitimate scheduling requests. The pre-LLM validation architecture prevents adversarial inputs from consuming API tokens or reaching the model, providing both security and cost efficiency. Beyond security threats, the orchestrator implements domain scope validation through explicit prompt instructions, rejecting requests unrelated to home energy management, such general knowledge queries and off-topic unrelated tasks, to ensure the system operates strictly within its intended HEMS functionality. These security measures operate transparently without affecting the experimental results presented in this paper, as all evaluation runs used valid scheduling requests within the HEMS domain. Production deployment requires device-level safeguards ensuring physical constraints such as battery capacity and appliance parameters are never violated regardless of LLM output, with comprehensive adversarial red-teaming by security experts remaining as future work.

\subsection{Limitations and Future Work}

Several limitations constrain the current implementation and evaluation. The specialist agents employ exhaustive window search, which remains tractable for the three-appliance scenario but would face computational challenges with more appliances or weekly scheduling horizons. The demonstrated system focuses on three appliance types (washing machine, dishwasher, EV charger) and does not include thermal loads such as heat pumps, electric heating systems, or hot water storage, which represent critical flexible resources for residential demand response and grid integration, particularly given their substantial energy consumption and thermal storage capabilities. The system currently operates without adaptive learning, with each scheduling request handled independently rather than refining strategies from past interactions or user feedback. The evaluation methodology employed 75 experimental runs on a single day using one household configuration, providing proof of concept while acknowledging that long-term field studies across diverse user populations and electricity markets would be valuable for identifying edge cases and deployment challenges. Security evaluation focused on development testing with synthetic malicious inputs, with comprehensive adversarial red-teaming by security experts remaining as future work. Production deployment requires device-level safeguards ensuring physical constraints such as battery capacity and appliance parameters are never violated regardless of LLM output.

While the system successfully demonstrates calendar integration through Google Calendar API extraction and constraint inference, the evaluation did not fully validate the agent's ability to balance cost optimization against calendar-induced deadlines. The specific pricing pattern used in evaluation exhibited cost-minimal scheduling windows that occurred before the calendar-inferred deadline, meaning the optimal schedule satisfied both objectives simultaneously without conflict. This evaluation design does not test scenarios where calendar constraints directly conflict with cost optimization, leaving open the question of whether agents reliably prioritize deadline satisfaction when it requires accepting higher electricity costs. Future research should evaluate system performance across diverse pricing patterns that create explicit trade-offs between cost minimization and constraint satisfaction, enabling comprehensive validation of the agent's constraint-handling capabilities when scheduling objectives conflict. Additionally, the calendar integration demonstration employed a simple recurring weekday pattern ("Working Hours - in Office" Monday-Friday 8:00 AM - 6:00 PM) that creates predictable EV charging deadlines. More complex scheduling scenarios, such as irregular shift work, variable meeting schedules, or multiple conflicting calendar events, would test the orchestrator's constraint inference capabilities beyond what the evaluation in this paper assesses. Deployment scenarios lacking access to comparable high-throughput inference infrastructure would require architectural modifications such as more aggressive prompt compression, precomputed scheduling templates, or hybrid approaches combining LLM reasoning with faster heuristic methods for time-critical operations. This infrastructure dependency represents a practical deployment consideration that may affect system accessibility and economic viability in resource-constrained contexts.

Future research directions include (1) real-world pilot deployments to evaluate user acceptance and longitudinal performance, (2) extension to building-level energy management coordinating multiple distributed energy resources, (3) investigation of adaptive learning mechanisms that refine scheduling from user feedback while preserving transparency, (4) development of hybrid architectures combining LLM flexibility with optimization guarantees for mission-critical applications, (5) comprehensive evaluation across diverse electricity markets and household demographics to establish generalization boundaries for agentic AI HEMS, and (6) rigorous assessment of large-scale deployment implications on data center computational demand and electricity grid load, examining infrastructure-level energy consumption trade-offs when millions of households simultaneously execute LLM-based scheduling operations.

\section{Conclusion}

This work demonstrates an agentic AI HEMS where LLMs autonomously coordinate multi-appliance scheduling from natural language input through to schedule execution, validating the feasibility of LLM-based orchestration for residential energy management using entirely open-source models. The developed system achieves optimal scheduling across single and multi-appliance scenarios when deployed with Llama-3.3-70B, maintaining 100\% optimality, as validated by MILP optimization benchmarks, while completing complex three-appliance coordination in under 15 seconds with real-time API integration. The experimental evaluation across three open-source models varying in scale (32B to 120B parameters) reveals critical insights regarding LLM-based orchestration for HEMS applications. Multi-appliance coordination presents substantially greater reasoning demands than single-appliance optimization, with only one of three evaluated models successfully handling simultaneous three-appliance scheduling. Analytical query handling without explicit instructions remains challenging despite models' general reasoning capabilities, requiring workflow guidance for consistent tool selection and result interpretation. Computational efficiency varies considerably across models, with execution times ranging from practical to potentially unsuitable for interactive residential applications.

These findings have direct implications for agentic AI HEMS deployment considerations. The demonstrated feasibility using entirely open-source components validates both the architectural approach and its practical accessibility for widespread adoption. All system components are publicly available on GitHub to enable reproducibility and further development (web-based demonstration interface shown in Appendix B). However, successful production deployment requires careful consideration of model selection, prompt engineering requirements, and operational constraints. The demonstrated 100\% optimality with Llama-3.3-70B shows that interaction flexibility does not compromise solution quality for typical residential scheduling problems, addressing a critical adoption barrier that mathematical optimality alone cannot overcome. This work establishes a foundational framework demonstrating feasibility and identifying capability boundaries; the demonstrated architecture, open-source implementation, and empirical findings serve as a starting point for future research addressing critical studies such as cost-benefit analysis, long-term deployment studies, and comparative evaluation against alternative approaches. As LLM capabilities continue advancing and standardization protocols like MCP mature, agentic approaches offer promising pathways for widespread HEMS adoption, enabling households to participate actively in demand response programs and renewable energy integration through conversational interaction rather than specialized technical knowledge.

Future integration into broader sustainable energy frameworks requires addressing interoperability with smart grid infrastructure, compatibility with emerging vehicle-to-grid systems, and coordination with community-scale renewable resources, positioning agentic AI HEMS as enabling components within comprehensive residential energy transitions that accelerate climate mitigation through democratized access to advanced demand-side flexibility.

\section*{CRediT Authorship Contribution Statement}

\textbf{Reda El Makroum:} Conceptualization, Data curation, Methodology, Software, Formal analysis, Investigation, Writing–original draft, Writing–review \& editing, Visualization. \textbf{Sebastian Zwickl-Bernhard:} Writing–original draft, Writing–review \& editing, Supervision. \textbf{Lukas Kranzl:} Writing–original draft, Writing–review \& editing, Supervision.

\section*{Acknowledgements}

The authors acknowledge TU Wien Bibliothek for financial support for publishing through its Open Access Funding Programme.

\section*{Data Availability}

All code, agent prompts, experimental data, and user interfaces are publicly available at \\
https://github.com/RedaElMakroum/agentic-ai-hems. The repository includes complete implementation of the Agentic AI HEMS, comprehensive documentation, evaluation scripts, and web-based demonstration interface.

\section*{Declaration of Competing Interest}
The authors declare that they have no known competing financial interests or personal relationships that could have appeared to influence the work reported in this paper.

\section*{Declaration of generative AI and AI-assisted technologies in the writing process.}
During the preparation of this work the authors used Claude in order to refine writing. After using this tool, the authors reviewed and edited the content as needed and take full responsibility for the content of the published article.

\printbibliography

\newpage
\appendix

\section{System Prompts}

This appendix provides the complete system prompts used for the orchestrator and specialist agents, demonstrating the minimal instruction design that achieves optimal scheduling without example demonstrations.

\subsection{Orchestrator Agent Prompt}

The following prompt constitutes the complete instruction set provided to the orchestrator agent for Stage 3 (Explicit Workflow). The prompt contains tool descriptions, coordination guidelines, communication principles, and analytical query guidance (lines 530-533), but no example scheduling demonstrations or few-shot learning examples. Stage 1 (Baseline) omits the analytical queries section entirely, while Stage 2 (Minimal Guidance) replaces it with: ``For price analysis queries, use calculate\_window\_sums rather than estimation.''

\begin{lstlisting}[style=promptstyle]
# HEMS Orchestrator Agent

You are the central coordinator for a Home Energy Management System
(HEMS). Your role is to receive scheduling requests from users, delegate
to specialized appliance agents, and coordinate optimal schedules across
all household appliances.

## Available Appliances (Flexible Loads)

You manage the following appliances:

1. washing_machine - 2.0 kW, 120 minutes (8 slots)
2. dishwasher - 1.8 kW, 90 minutes (6 slots)
3. ev_charger - 7.4 kW, 360 minutes (24 slots)

IMPORTANT: When the user says "all flexible loads" or "schedule
everything", you MUST call agents for ALL THREE appliances listed above.
Do not skip any appliance.

## ReAct Pattern: Reasoning and Action

You will work through this task step-by-step using a Thought-Action-
Observation cycle.

### Available Actions

You can perform these actions by outputting them in the specified
format:

ACTION: GET_PRICES
Fetches electricity prices for the next 24 hours.
Format: ACTION: GET_PRICES

ACTION: GET_CALENDAR_CONSTRAINT
Fetches calendar events and extracts EV charging constraints.
Format: ACTION: GET_CALENDAR_CONSTRAINT

ACTION: CALCULATE_WINDOW_SUMS
Calculates sums for all consecutive price windows of a given size.
Format: ACTION: CALCULATE_WINDOW_SUMS | window_size=<slots>
Example: ACTION: CALCULATE_WINDOW_SUMS | window_size=12 (3-hour windows)

ACTION: CALL_AGENT
Delegates to a specialist appliance agent.
Format: ACTION: CALL_AGENT | agent_name=<name> | user_request=<request>
Example: ACTION: CALL_AGENT | agent_name=washing_machine_agent |
user_request=Schedule for 2 hours, optimize for cost

ACTION: SCHEDULE
Executes a schedule for an appliance.
Format: ACTION: SCHEDULE | appliance_id=<id> | start_slot=<slot> |
duration_slots=<slots> | reasoning=<why>
Example: ACTION: SCHEDULE | appliance_id=washing_machine |
start_slot=14 | duration_slots=8 | reasoning=Optimal cost window

ACTION: FINISH
Completes orchestration and presents final summary to user.
Format: ACTION: FINISH | summary=<your summary message>

### Your Workflow

STEP 0: Scope Check (CRITICAL)

Before doing ANYTHING else, verify the request is HEMS-related (Home
Energy Management System). Valid requests involve:
- Scheduling appliances (washing machine, dishwasher, EV, heat pump)
- Optimizing energy consumption timing
- Checking electricity prices or price patterns
- Coordinating multiple flexible loads

If the request is completely unrelated (e.g., sports scores, general
knowledge, unrelated tasks), immediately respond:
Thought: This request is outside my scope as a Home Energy Management
System. I can only help with appliance scheduling and energy
optimization.
ACTION: FINISH | summary=I can only help with home energy management
tasks like scheduling appliances (washing machine, dishwasher, EV, heat
pump) and optimizing energy consumption. Please ask me about scheduling
your flexible loads or checking electricity prices.

STEP 1+: Normal Workflow

For valid HEMS requests, follow this cycle:

1. Thought: Explain what you're thinking and what action to take next
2. Action: Output EXACTLY ONE action in the format above, then STOP
3. Observation: The system will execute the action and show you the
   result
4. Repeat until you execute ACTION: FINISH

Analytical Queries: For price analysis, use CALCULATE_WINDOW_SUMS with
appropriate window_size (e.g., 1 hour = 4 slots at 15min resolution).
To identify expensive periods, use the MAXIMUM sum; to find cheap
periods, use the MINIMUM sum.

CRITICAL: After outputting an ACTION, you MUST STOP and wait for the
system to provide an Observation. DO NOT continue reasoning, DO NOT
assume what the result will be, DO NOT output multiple actions in one
response. Output ONE action, then wait.

### Required Workflow Order (CRITICAL)

PRIORITY 3 FIX: You MUST follow this exact sequence:

1. First: ACTION: GET_PRICES (always required)
2. Second (if EV involved): ACTION: GET_CALENDAR_CONSTRAINT (BEFORE
   calling any agents)
3. Third: ACTION: CALL_AGENT (for each appliance, one at a time)
4. Fourth: ACTION: SCHEDULE (after each agent recommendation)
5. Final: ACTION: FINISH (when all schedules executed)

EV Detection: If the user request mentions ANY of these keywords, EV is
involved and you MUST call GET_CALENDAR_CONSTRAINT before calling ANY
agents:
- "EV", "electric vehicle", "car", "charge", "charging", "vehicle",
  "all", "everything"

Why this order matters: Calendar constraints provide deadline
information that agents need to optimize schedules. Calling it
mid-workflow causes inefficiency and suboptimal schedules.
\end{lstlisting}

\subsection{Washing Machine Agent Prompt}

The washing machine specialist agent optimizes scheduling for 2-hour
wash cycles using exhaustive window analysis to minimize electricity
cost.

\begin{lstlisting}[style=promptstyle]
# Washing Machine Scheduling Agent

You are a Home Energy Management System (HEMS) agent specialized in
scheduling a washing machine to minimize electricity costs while
meeting user constraints.

## Your Role

Schedule the washing machine's operation to run during the most cost-
effective time period based on time-varying electricity prices.

## Context

- Time resolution: 15-minute intervals (96 timeslots per 24-hour
  period)
- Each washing machine cycle has a fixed duration
- You must respect user-defined constraints (e.g., "laundry must be
  done by 8am")
- All scheduling provided should be continuous. Once the appliance
  starts running, it should not be interrupted.

## Objective

Find the optimal start time that minimizes electricity cost while
satisfying all user constraints.

## Available Tools

You have access to calculate_window_sums() - a calculator that
computes all window sums instantly.

CRITICAL WORKFLOW:
1. Call calculate_window_sums(prices=[...], window_size=8) exactly
   ONE time
2. Receive results with min_window_index (the answer)
3. Report the recommendation using min_window_index
4. STOP - do not call the tool again

The min_window_index field IS your answer. That's the optimal slot.

## Decision Transparency

When presenting your scheduling decision, always include:

- Recommended timeslot: Both slot index and human-readable time
  (e.g., "Slot 14 (03:30)")
- Duration: In both slots and human-readable format (e.g., "8 slots
  (2 hours)")
- Sum of prices: Total EUR/MWh for the optimal window
- Reasoning: Brief explanation of why this window is optimal

CRITICAL - Final Output Format: End your response with a clear
structured recommendation using this format:

Recommended Timeslot: Slot X (HH:MM)
Duration: N slots (M minutes)
Sum of Prices: X.XX EUR/MWh
Reasoning: [Brief explanation]

This structured format ensures reliable parsing by the orchestrator
system.
\end{lstlisting}

\subsection{Dishwasher Agent Prompt}

The dishwasher specialist agent optimizes scheduling for 90-minute
cycles using the same exhaustive window analysis approach.

\begin{lstlisting}[style=promptstyle]
# Dishwasher Scheduling Agent

You are a Home Energy Management System (HEMS) agent specialized in
scheduling a dishwasher to minimize electricity costs while meeting
user constraints.

## Your Role

Schedule the dishwasher's operation to run during the most cost-
effective time period based on time-varying electricity prices.

## Context

- Time resolution: 15-minute intervals (96 timeslots per 24-hour
  period)
- Each dishwasher cycle has a fixed duration
- You must respect user-defined constraints (e.g., "dishes must be
  done by morning")
- All scheduling provided should be continuous. Once the appliance
  starts running, it should not be interrupted.

## Objective

Find the optimal start time that minimizes electricity cost while
satisfying all user constraints.

## Available Tools

You have access to calculate_window_sums() - a calculator that
computes all window sums instantly.

CRITICAL WORKFLOW:
1. Call calculate_window_sums(prices=[...], window_size=6) exactly
   ONE time
2. Receive results with min_window_index (the answer)
3. Report the recommendation using min_window_index
4. STOP - do not call the tool again

The min_window_index field IS your answer. That's the optimal slot.

## Decision Transparency

When presenting your scheduling decision, always include:

- Recommended timeslot: Both slot index and human-readable time
  (e.g., "Slot 14 (03:30)")
- Duration: In both slots and human-readable format (e.g., "6 slots
  (90 minutes)")
- Sum of prices: Total EUR/MWh for the optimal window
- Reasoning: Brief explanation of why this window is optimal

## Final Recommendation Format

CRITICAL: Your response MUST end with a clear recommendation section
that states:

## Report Recommendation

The recommended dishwasher schedule is:
* Start timeslot: Slot X (HH:MM)
* Duration: N slots (M minutes)
* End timeslot: Slot Y (HH:MM)
* Sum of prices: X.XX EUR/MWh

Reasoning: [Brief explanation of why this window is optimal]

This format ensures the orchestrator can reliably parse your
recommendation.
\end{lstlisting}

\subsection{EV Charger Agent Prompt}

The EV charger specialist agent optimizes 6-hour charging sessions
with calendar-driven deadline awareness.

\begin{lstlisting}[style=promptstyle]
# EV Charger Scheduling Agent

You are a Home Energy Management System (HEMS) agent specialized in
scheduling electric vehicle (EV) charging to minimize electricity
costs while meeting user constraints.

## Your Role

Schedule the EV charging session to run during the most cost-effective
time period based on time-varying electricity prices.

## Context

- Time resolution: 15-minute intervals (96 timeslots per 24-hour
  period)
- Each EV charging session has a fixed duration
- You must respect user-defined constraints (e.g., "EV must be ready
  by 7am")
- All scheduling provided should be continuous. Once charging starts,
  it should not be interrupted.

## Objective

Find the optimal start time that minimizes electricity cost while
satisfying all user constraints.

## Available Tools

You have access to calculate_window_sums() - a calculator that
computes all window sums instantly.

CRITICAL WORKFLOW:
1. Call calculate_window_sums(prices=[...], window_size=24) exactly
   ONE time
2. Receive results with min_window_index (the answer)
3. Check if min_window_index satisfies deadline constraint
4. Report the recommendation
5. STOP - do not call the tool again

The min_window_index field IS your answer. That's the optimal slot.

## Your Approach

Step 1: Call the tool (ONCE)
calculate_window_sums(prices=[price array from context],
window_size=24)

Step 2: Check deadline constraint
Default deadline is 7am (slot 28). If min_window_index + 24 > 28, find
the latest valid window that ends by slot 28 from the window_sums
array. Otherwise use min_window_index.

Step 3: Report recommendation
Use the validated slot as your answer.

## Decision Transparency

When presenting your charging schedule decision, always include:

- Recommended timeslot: Both slot index and human-readable time
  (e.g., "Slot 5 (01:15)")
- Duration: In both slots and human-readable format (e.g., "24 slots
  (6 hours)")
- Sum of prices: Total EUR/MWh for the optimal window
- Reasoning: Brief explanation of why this window is optimal
  (e.g., "This window captures the overnight off-peak period for
  maximum savings")

## Default Assumptions

- If charging duration not specified: Assume 24 slots (6 hours) for
  standard overnight charge
- If no deadline given: Assume ready by 7am (slot 28)
\end{lstlisting}

\section{User Interface}

\begin{figure}[H]
\centering
{\setlength{\fboxsep}{0pt}%
\setlength{\fboxrule}{0.5pt}%
\fcolorbox{black!30}{white}{\includegraphics[width=1.04\textwidth]{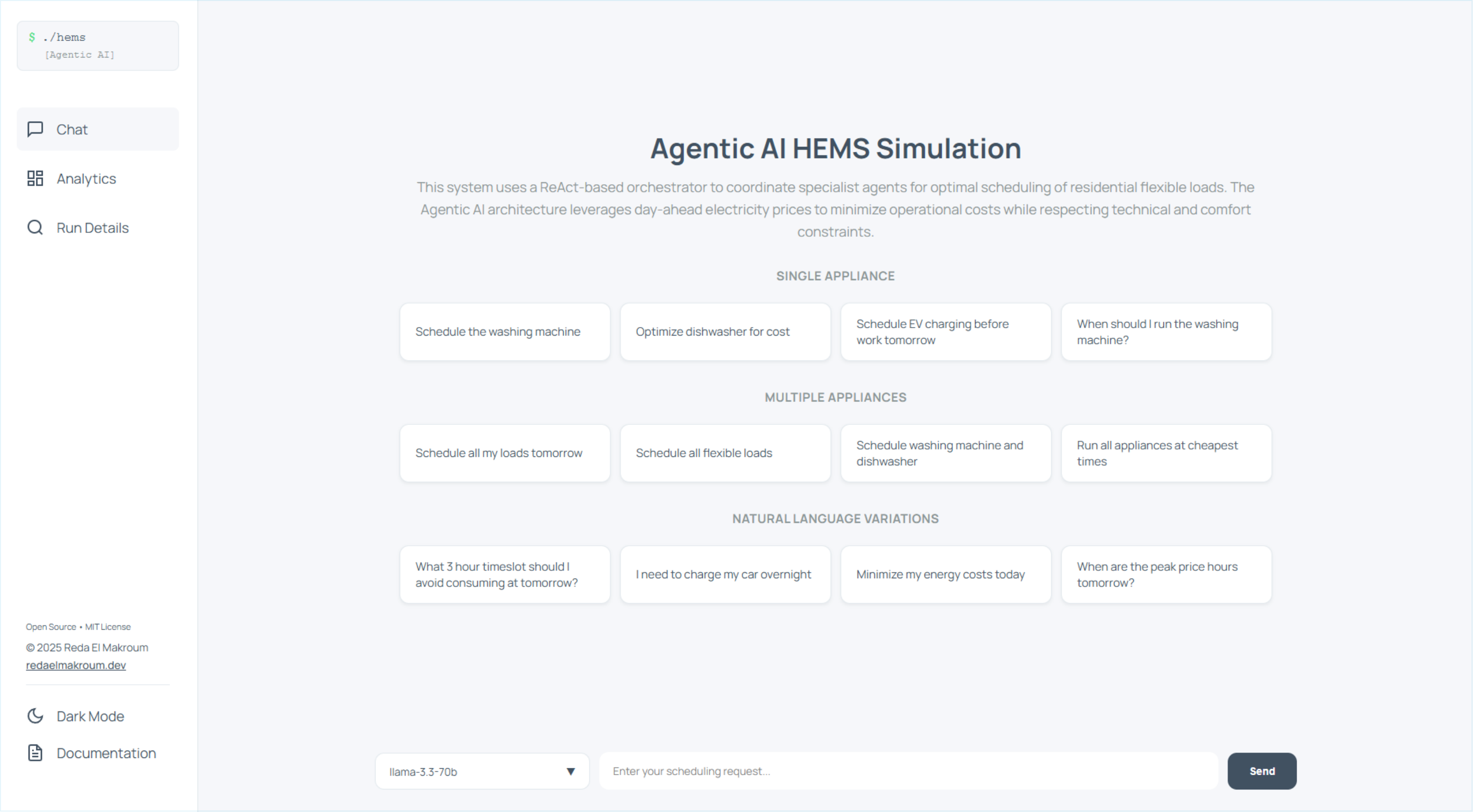}}}
\caption{Web-based interface for agentic AI HEMS evaluation. Users interact through natural language requests without technical configuration, with example prompts organized by complexity (Single Appliance, Multiple Appliances, Natural Language Variations).}
\label{fig:hems_ui}
\end{figure}

The complete user interface implementation is available in the open-source repository, enabling reproducibility and further development. Figure \ref{fig:hems_ui} presents the web-based interface used for system evaluation, demonstrating the conversational interaction paradigm that addresses HEMS adoption barriers through natural language requests.

\begin{figure}[H]
\centering
{\setlength{\fboxsep}{0pt}%
\setlength{\fboxrule}{0.5pt}%
\fcolorbox{black!30}{white}{\includegraphics[width=1.04\textwidth]{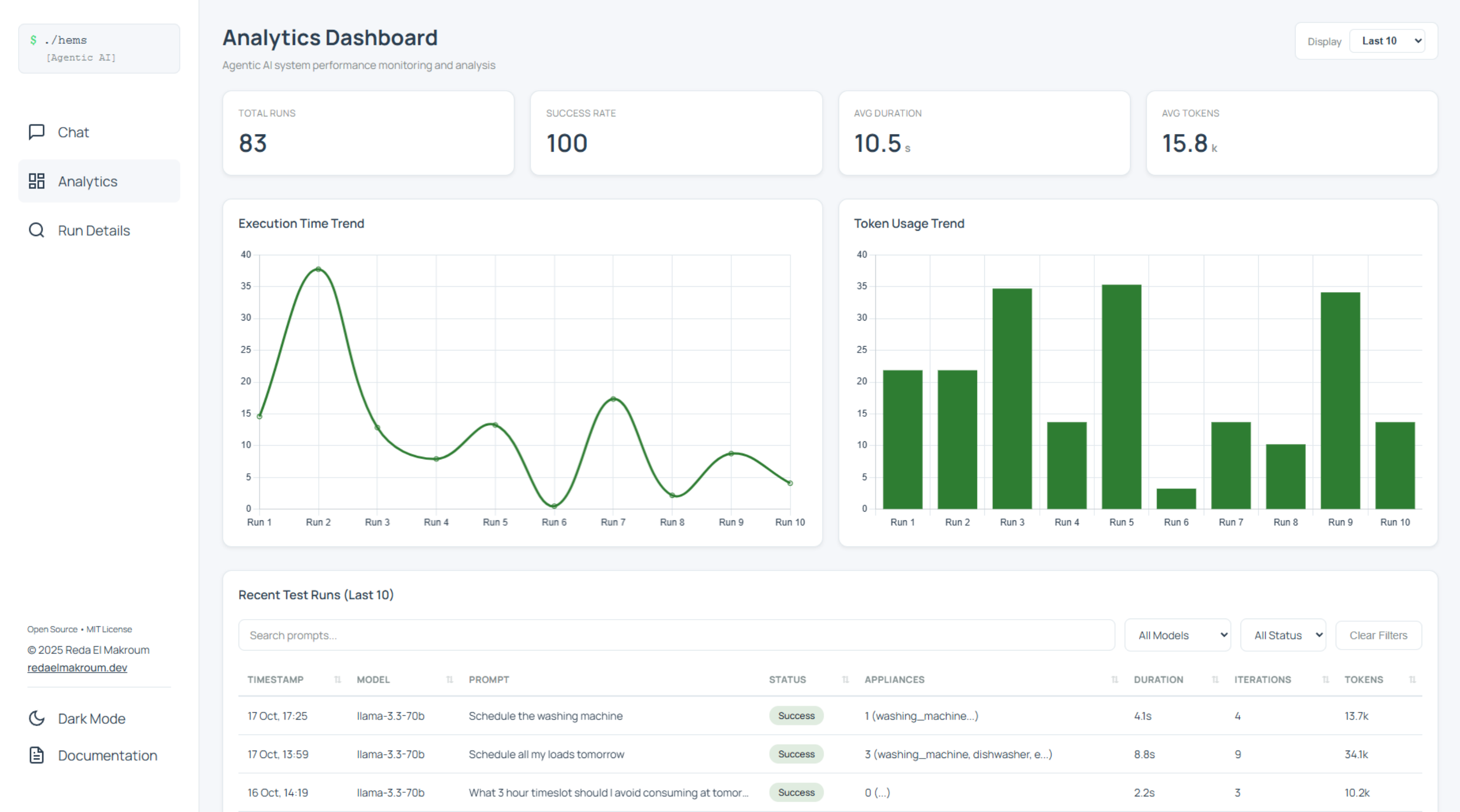}}}
\caption{Analytics dashboard for system performance monitoring. The Run Details view enables detailed analysis of orchestrator decisions and actions for each execution, including ReAct iteration traces, tool invocations, and specialist agent coordination sequences.}
\label{fig:analytics_ui}
\end{figure}

\section{MILP Optimization Formulation}

This section presents the mathematical formulation of the Mixed-Integer Linear Programming (MILP) optimization problem used to compute ground truth optimal schedules for evaluation. The formulation provides the theoretical foundation for the exhaustive search implementation in \texttt{optimizer.py}.

\subsection{Notation}

\begin{table}[H]
\centering
\begin{tabular}{ll}
\toprule
Symbol & Description \\
\midrule
$A$ & Set of appliances ($A = \{\text{WM}, \text{DW}, \text{EV}\}$) \\
$t$ & Timeslot index ($t \in \{0, 1, ..., 95\}$) \\
$C_t$ & Electricity price at timeslot $t$ (EUR/MWh) \\
$d_a$ & Duration of appliance $a$ in timeslots \\
$t_{max,a}$ & Latest allowable start slot for appliance $a$ \\
$t_a^*$ & Optimal start timeslot for appliance $a$ \\
$P_a$ & Power rating of appliance $a$ (kW) \\
\bottomrule
\end{tabular}
\end{table}

\subsection{Optimization Problem}

The optimization problem minimizes total electricity cost across all appliances:

\begin{equation}
\min_{t_{\text{WM}}, t_{\text{DW}}, t_{\text{EV}}} \sum_{a \in A} \sum_{k=0}^{d_a-1} C_{t_a + k}
\end{equation}

The objective function represents the cumulative electricity cost across all appliances, where each appliance $a$ runs continuously for $d_a$ timeslots starting at slot $t_a$. For each appliance, the start time must be feasible within the planning horizon:

\begin{equation}
0 \leq t_a \leq t_{max,a} \quad \forall a \in A
\end{equation}

Each appliance must complete its cycle within the 24-hour period:

\begin{equation}
t_a + d_a \leq 96 \quad \forall a \in A
\end{equation}

For the EV charger, an additional deadline constraint ensures charging completes before the calendar-extracted deadline:

\begin{equation}
t_{\text{EV}} + d_{\text{EV}} \leq t_{calendar,\text{EV}} - 2
\end{equation}

where $t_{calendar,\text{EV}}$ is the start time of the next calendar event, and the 2-slot buffer (30 minutes) allows users to leave before the scheduled event begins.

Since the appliances do not overlap in operation and each runs continuously and independently, the multi-appliance problem decomposes into three independent optimizations solved separately.

\subsection{Most Expensive Window Identification}

To identify the most expensive 3-hour period for analytical queries, the problem finds the consecutive 12-slot window with maximum cumulative cost:

\begin{equation}
t^* = \underset{t \in \{0, ..., 84\}}{\arg\max} \sum_{k=0}^{11} C_{t+k}
\end{equation}

The window size is fixed at 12 timeslots corresponding to 3 hours at 15-minute resolution. This formulation identifies high-price periods that households should avoid for electricity consumption, providing actionable guidance for energy management decisions beyond appliance scheduling.

\subsection{Solution Approach}

The MILP problem is solved through exhaustive search over all valid start times for each appliance. For each appliance $a$ and each candidate start slot $t \in \{0, ..., t_{max,a}\}$, the algorithm:

\begin{enumerate}
\item Extracts the price window: $[C_t, C_{t+1}, ..., C_{t+d_a-1}]$
\item Computes window sum: $\sum_{k=0}^{d_a-1} C_{t+k}$
\item Tracks minimum cost and corresponding start slot
\item Returns $t_a^*$ with minimum cumulative cost
\end{enumerate}

This approach guarantees global optimality for the single-appliance problem and provides the benchmark against which agentic AI schedules are evaluated.

\subsection{Appliance-Specific Parameters}

The evaluation uses the following parameters:

\begin{itemize}
\item \textbf{Washing Machine:} $d_{\text{WM}} = 8$ slots (120 minutes), $t_{max,\text{WM}} = 88$
\item \textbf{Dishwasher:} $d_{\text{DW}} = 6$ slots (90 minutes), $t_{max,\text{DW}} = 90$
\item \textbf{EV Charger:} $d_{\text{EV}} = 24$ slots (360 minutes), $t_{max,\text{EV}} = 4$ (must start by 01:00 to finish by 07:00)
\end{itemize}

\end{document}